\documentclass[11pt,a4paper]{article}
\usepackage[hyperref]{emnlp-ijcnlp-2019}
\usepackage{times}
\usepackage{latexsym}
\usepackage{graphicx}
\usepackage{url}
\usepackage{algorithmic}
\usepackage{lipsum}
\usepackage{todonotes}
\usepackage{float}
\usepackage{subfig}
\usepackage{enumitem}

\aclfinalcopy 

\title{Rating for Parents: Predicting Children Suitability Rating for Movies Based on Language of the Movies}

\author{Mahsa Shafaei, Niloofar Safi Samghabadi, Sudipta Kar,  Thamar Solorio \\
  Department of Computer Science, University of Houston  \\
  {\tt \{mshafaei, nsafisamghabadi, skar3\}@uh.edu}, \tt {solorio@cs.uh.edu}}

\begin{document}
\maketitle
\begin{abstract}
    The film culture has grown tremendously in recent years. The large number of streaming services put films as one of the most convenient forms of entertainment in today's world. Films can help us learn and inspire societal change. But they can also negatively affect viewers.
    In this paper, our goal is to predict the suitability of the movie content for children and young adults based on scripts. The criterion that we use to measure suitability is the MPAA rating that is specifically designed for this purpose. We propose an RNN based architecture  with attention that jointly models the genre and the emotions in the script to predict the MPAA rating. We achieve 78\% weighted F1-score for the classification model that outperforms the traditional machine learning method by 6\%. 
\end{abstract}

\section{Introduction}

The Motion Picture Association of America (MPAA) is a film rating system that establishes the appropriate age for movie viewers \footnote{\url{https://www.mpaa.org/film-ratings/}}. The MPAA rating is determined by CARA\footnote{Classification \& Ratings Administration} (one of the subdivisions of MPAA organization). Members of CARA watch the entire film to identify the age category and the MPAA rating of the movie \cite{motion2010classification}. The current method for rating movies has three major issues. First, it is a time-consuming process that is not scalable. Second, there are many unrated online movies on the Internet. And, there is no automatic system for users like parents and media service providers to specify the age category of movies. Finally, it needs to be done after production, when making changes in movies can cost a lot of money. In this paper, we want to automatically predict this rating at early steps when we only have the script of the movie.


Many research has been shown that movies can affect people's behavior directly, and young viewers may be especially open to being influenced ~\cite{sargent2006alcohol, american2001media} . MPAA ratings have a wide practical value. For example, parents can rely on them as a guideline to determine what movies can their children watch. Also, media service providers (e.g. Amazon and Netflix) rely on these ratings to enable age filters in parental controls.
Moreover, having the MPAA rating is an important element for producers too. Although films can be exhibited without a rating, certain theaters refuse to exhibit non-rated movies\footnote{\url{https://en.wikipedia.org/wiki/\\Motion_Picture_Association_of_America_film_rating_system}}, and it certainly can affect gross revenue of movies. By automatically predicting the MPAA rating, we can achieve these outcomes:
First, it is faster than the manual version, so it is applicable to a large number of movies. Second, movies can be rated before the production is complete, and producers can use the predicted rating to make adjustments based on their audience at early stages and save money.  Finally, using this system, users can find the rating for many movies out there that are still unrated.

Predicting the MPAA rating (based on the script of the movie) is not a trivial task. Having a list of bad words is not always sufficient to predict the level of suitability of movies for different ages since swearing impact is a function of communication context as shown in \cite{jay2008pragmatics}. For example, the impact depends on the social position of the speaker e.g. a student using these words is not as bad as a principal using them.  So, we need to analyze these words in their context to be able to indicate the level of bad influence. The use of a bad word list as a single source is also inefficient since the MPAA rating is not dedicated to a specific aspect like offensive language  \cite{jenkins2005evaluation}. It is a combination of several elements like drugs, sex, violence, language, thematic material, adult activities, etc. Moreover, current sources for bad word lists are not designed to cover all these aspects. Similarly, based on \cite{jenkins2005evaluation} and \cite{webb2007violent} violence prediction is not enough to predict the MPAA rating either. There are some previous works on violence prediction in movies \cite{martinez2019violence}, but to the best of our knowledge, none of them aimed at predicting the MPAA rating. So, previous works cannot cover our goal.

Our paper is the first paper that works on predicting the MPAA rating. We introduce a novel deep learning model to automatically solve the problem. The devised deep learning architecture jointly models conversations between characters, genres of the movie, and emotions in the conversations to predict MPAA ratings. Our proposed architecture is an RNN-based model with an attention layer. Also, in our proposal we let the model learn everything from the raw data (end-to-end learning), and we avoid the use of lexicons of bad words because the list of bad words could change over the time. Our system only exploits the textual information via the scripts with the idea of being able to predict rating at early steps of production when we only have the script. However, in our future work we are also interested in exploiting audio and video information. This paper presents the following main contributions:

\begin{itemize}
    \setlength{\itemsep}{1pt}
    \item A novel task: automatic prediction of the suitability of movies for children using the MPAA rating scheme. 
    
    \item The creation of the first corpus of movie scripts with their associated MPAA rating. This dataset contains $\approx${7K} movies from 24 different genres.
    
    \item A deep learning model for predicting the MPAA rating with 78\% weighted F1-score that works 6\% better than the traditional machine learning method.
    
    \item A large-scale empirical study to show the effect of genre of the movie and emotion in the conversation on suitability prediction model.

\end{itemize}


\section{Related Work}
 Although to the best of our knowledge there is no previous work on predicting the MPAA rating, there are existing related works on detection of violent content, abusive language, and hate speech. 
 We categorize previous works into two groups based on their underlying approach: first, papers that present a traditional Machine Learning model, and the second group works with Deep Neural Network models.
\subsection{Traditional Approach}
Researchers in \cite{giannakopoulos2010audio}  predicted violence in movies by extracting visual and audio features from movies. Authors in \cite{gninkoun2011automatic} built upon that and added textual features to visual and audio features to capture inappropriate words for violence detection.  We also use a bad word list to create a baseline model to compare with our proposed models.  
On the other hand, based on the survey done on hate speech detection \cite{schmidt2017survey}, several works have been done on detecting offensive language in text. These works are related to our work because offensive language in dialogues can affect suitability of movies for children. For example, authors of \cite{davidson2017automated} extract lexical, syntactic and sentiment features and trained a multi-class classifier to automatically predict hate speech on Twitter data. In another work \cite{nobata2016abusive}, authors used the same type of features along with a SVM classifier to build an abusive language detection system in online user content.
For our traditional model, we also extract lexical and sentiment features and feed them into a SVM classifier.
\subsection{Deep Learning Approach}
Employing the language of movies, authors of \cite{martinez2019violence} tried to predict if a movie is violent or not. This paper is the closest paper to our work because they only use dialogues of the movies as well. They extracted sentiment, semantic and lexical features and employed SVM and RNN-based classification models to predict violence in movies based on the extracted features. However, in our work, we use raw data (conversation between characters) to predict MPAA ratings using our deep learning model. Convolution and Recurrent Neural network have been used to predict abusive language and hate speech on Twitter data
\cite{singh2018aggression,zhang2018detecting,park2017one,mathur2018did}, but there is no similar works on movie corpus.

\section{Dataset}
We employ the movie script dataset collected by \cite{mshafaei}. They provided the scripts of about 15K movies as well as their metadata like name of actors, directors, genre, and etc.  It should bee noted that the scripts contain only conversation between characters. The dataset does not include MPAA rating, so we crawl the IMDB website to extract MPAA rating. Our final dataset contains about 7K movies since not all the movies in the original dataset have the MPAA rating. There are five categories for the MPAA rating (G, PG, PG-13, R, NC-17) that specify the suitability of movies for children. G stands for general group, it means all ages admitted. PG means there are some materials in the movie that parents need to check. PG-13 indicates some materials in the movie are not appropriate for children under 13 years old. R stands for restricted and means people under 17 need to watch the movie with a parent. NC-17 refers to no one under 17 is allowed to watch the movie. The exact definition of these rating are available at (https://www.mpaa.org).

Table \ref{table:Data_Statistics} shows the statistics of different categories in our dataset. Based on this table, we have an imbalanced dataset that includes very few numbers of NC-17 and G rated movies.

\begin{table}[ht]
\centering
\resizebox{.95\columnwidth}{!}{
\begin{tabular}{|c|c|c|c|c|c|c|c|}
\hline
\multicolumn{1}{|l|} {{\textbf{Rating}}}& {\textbf{G}} &{\textbf{PG}}& {\textbf{PG13}}  & {\textbf{R}} &{\textbf{NC17}}&{\textbf{Total}} \\
\hline
\multicolumn{1}{|l|} {\#Movies} &92 &1056   &1862   &4030   &14  &7040           \\
\hline

\end{tabular}
}

\caption{\label{table:Data_Statistics} Dataset statistics; number of movies in each category of MPAA rating. G is the most proper and NC-17 is the least appropriate group for children.}
\end{table}

In Table \ref{table:genre_dis}, we show the distribution of movies in different genres. Class imbalance exists between different genres since some genres are more popular in the move industry. We decided not to fix this issue to keep the dataset representative of the real world situation.

\begin{table}[ht]
\centering
\footnotesize
\begin{tabular}{|l|r||l|r|}
\hline
\textbf{Genre} &\textbf{\#}& \textbf{Genre} &\textbf{\#}\\ \hline
{Science-Fiction}  &502   & {Action} &1550              \\
{Horror}  &912   & {Animation} &283                  \\
{Crime} &1303  & {Adventure} &964         \\
{Romance} &1149  &  {History}  &211        \\
{News} &5     & {Western}  &78       \\
{Comedy} &2427  & {War}  &160                \\   
{Thriller}  &1320   & {Short}  &17          \\
{Mystery}  &593  & {Film-Noir}  &15    \\
{Musical}  &260   & {Drama}  &3693       \\
{Documentary}  &192   & {Family}  &393      \\
{Sport}  &191 & {Biography}  &458     \\
{Fantasy}  &457   &   &    \\
\hline

\end{tabular}
\caption{\label{table:genre_dis} Distribution of movies in each genre (some movies are assigned to more than one genre).}
\end{table}

Our dataset will be available to public to encourage people to research on this area.

\section{Methodology}
\begin{figure}[h]
\centering
\includegraphics[width=\columnwidth]{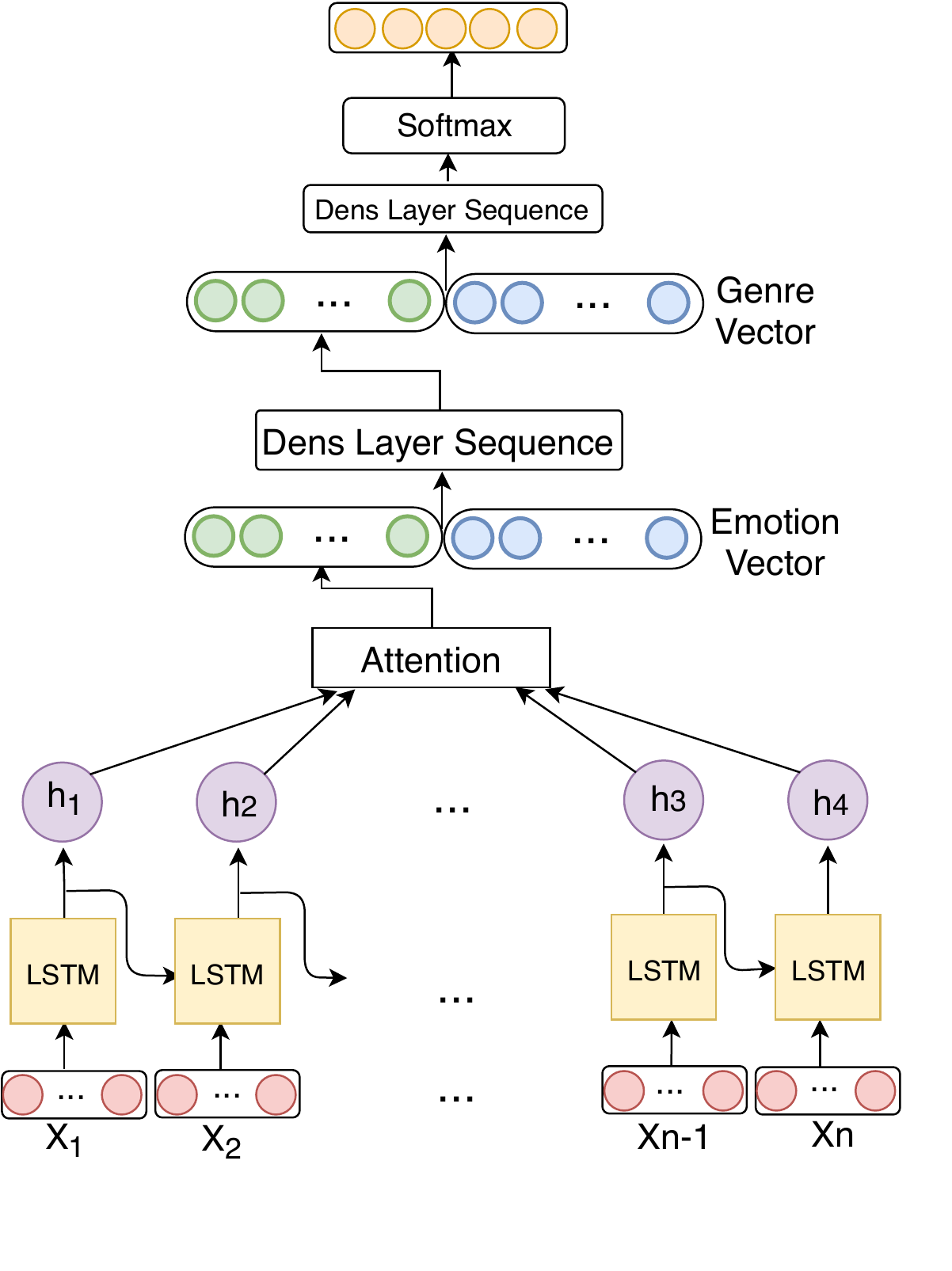}

\caption{Overall architecture of the proposed model}
\label{fig:arch}
\end{figure}

The main goal of the paper is to classify movies into one of the MPAA ratings based on the content of the movie. To achieve this goal, we use three types of resources in our model; raw data (conversations between characters), emotion vectors of these conversations, and genre vectors of the movies.
 Figure \ref{fig:arch} shows the architecture of the neural network model that we use. The model consists of 1) an embedding Layer to convert the words into the vector representation  2) a Long short-term memory (LSTM) layer to learn the spatial dependency of the words 3) an attention layer to assign weights to words in the sequence, 4) emotion and genre vectors to add contextual information to the model, and 5) dense and softmax layers to predict the rating.  
In the following sections, we explain each layer of the model. 

\subsection{Embedding Layer}
Word embedding is an effective technique for text classification as it is capable to capture semantic information of the text.
The input of the model is an embedding layer that gets a vector of words' indexes $[I_1,I_2,...,I_{10000}]$, and the output of the layer is a 2-D matrix  $[[v_{1 ,1},...,v_{1,j}],[v_{2,1},...,v_{2,j}],...,[v_{n,1},...,v_{n,j}]]$; each vector $[v_{i ,1},...,v_{i,j}]$ is the embedding vector for the corresponding word $i$. We use 300 dimensional pre-trained Glove embedding\footnote{\url{https://nlp.stanford.edu/projects/glove/}}.

\subsection{LSTM Layer with an Attention}
To extract the sequential information from the scripts, we use a LSTM layer. This layer transforms a sequence of embedded vectors to a sequence of hidden vectors. Then we pass the resulting hidden representation to the attention mechanism. We use the same attention model proposed in \cite{bahdanau2014neural}. This layer computes the weighted sum $r$ as $\sum_i\alpha_ih_i$ to aggregate hidden layers of LSTM to a single vector. The model can learn the relative importance of hidden states by learning the  $\alpha_i$.
 We compute $\alpha_i$  as follow:
\begin{equation}
\alpha_i = \mbox{softmax}(v^T \mbox{tanh}(W_h h_i + b_h)) \\
\end{equation}
where $h_i$ is  the hidden state and  $v$ and $W_h$ are parameters of the network.
\subsection{Emotion Vector}
Emotion of the conversations in the movie can help the model to capture the context of the words; also dominant emotions of the movie can be a factor that shows the suitability of the movie for children e.g. fear or disgust. So, we concatenate the emotion vector with the attention output.
To extract emotion of the text, we use NRC emotion lexicon \cite{mohammad2011once}. This dictionary maps words to eight different emotions (anger, anticipation, joy, trust, disgust, sadness, surprise, and fear) and two sentiments (positive and negative) with binary values. we calculate the normalized count of words per emotion over the whole movie. As a result, we have a vector $[e_{1},e_2,...,e_{10}]$ for each movie, where $e_i$ is the percentage of words corresponding to emotion $i$.

\subsection{Genre Vector}
Genre can provide information about the theme of the movie. For example, some dialogues are considered as violent in some specific genre, but they are harmless in another genre \cite{martinez2019violence}. So, we consider genre by adding it to the model as a vector. We have a total of 24 genres across our corpus, but some of the movies are assigned in several genres. Thus, we form a multi-hot vector for modeling the genres of a movie; ones stands for genres of the movie and zeros for other genres.

\subsection{Dense Layer and Output Layers}
We have two dense layer sequences in our model.  We use batch normalization and dropout after the hidden layer to avoid over-fitting and achieve a more generalized model. Since we have a multi-class classification, in the final step, we use the softmax activation function to calculate the probability of each group (G, PG, PG-13, R, NC-17). 

\section{Experiments}

As our data is imbalanced, we use the random stratified sampling to split data to 80:10:10 training, development, and test set (for all experiment we use the same train, validation and test data).  We also choose weighted F1-score as the evaluation metric.

\subsection{Baseline Systems}
There is no previous study available of the same topic, so we define three baselines to evaluate the performance of our work; one optimization method, one traditional machine learning method, and one deep model. 

\noindent\textbf{Threshold Model:} Our first baseline only considers the bad words that have been used in the movie scripts. It finds the best threshold for the percentage of bad words to classify movies. It labels movies with the lowest number of bad words as G-rated and those with highest number of bad words as NC-17. To create our bad words list, we compiled a list from Google's bad words list~\footnote{\url{https://code.google.com/p/badwordslist/downloads/detail?name=badwords.txt}} and words listed in ~\cite{hosseinmardi2014towards}. Using this list, we calculate the percentage of bad words in each movie. Then, for each rating group (G, PG, PG-13, R, and NRC-17), we define a threshold $t_i$ and train the model on training set to find these thresholds. The final model will be a list of thresholds: all movies with bad words less than $t_1$ are labeled as G, all movies with bad words between $t_1$ and $t_2$ are labeled as PG, etc. The intuition behind this baseline is to show that having bad words are not enough to decide about suitability of movies for children.

\noindent\textbf{SVM Model:} The second baseline model is similar to the model proposed for movie success prediction task \cite{mshafaei}. The best set of features for the model is a combination of unigram, bigram, bag-of-genres and bag-of-directors. We also add emotion vector to the feature set to have a fare comparison with our deep learning model. Using the grid search method, we tune hyper-parameter C of the SVM model, $C \in\{1, 10, 100, 1000\}$.

\noindent\textbf{CNN Model:}  The third baseline model is a strong deep neural model which
has the same architecture as our proposed model, but it uses Convolutional Neural Network (CNN) instead of LSTM layer with an attention. CNN is used previously for detecting offensive language and hate speech ~\cite{aroyehun2018aggression,zhang2018detecting}.   



\subsection{Experimental Setup}
We use Pytorch to implement our RNN model. 
To tune hyper parameters, we run experiments on validation set for the model with different values of learning rate \{0.00001, 0.0001\}, number of LSTM's hidden units \{32, 64, 128, 256\}, and dropout rates \{0.3, 0.4, 0.5\}. Also, to avoid over-fitting, in addition to dropout, we use L2 regularization. We use binary cross entropy loss function in order to calculate the loss between predicted and actual labels and employ Adam as the optimizer \cite{kingma2014adam}. Best performance obtained by  (dropout = 0.3, learning rate = 0.00001, LSTM hidden units = 256). We train over 200 iterations and consider the model with the best weighted-F1 score on the validation set as a the final model to apply on the test set. 
  
\section{Results}

\begingroup
\begin{table}[ht]
\centering
\footnotesize
\begin{tabular}{|c|c|}
\hline
\multicolumn{1}{|c|} { Models} & {F1-Score}\\ \hline
\multicolumn{1}{|l|} {{Baseline 1- Threshold model}}  &63.90                 \\
\multicolumn{1}{|l|}{{Baseline 2- SVM}}  &71.93                 \\
\multicolumn{1}{|l|}{{Baseline 3- CNN}}  &74.74                 \\\hline
\multicolumn{1}{|l|}{{LSTM with Attention layer (L\&A)}}  &75.80                 \\
\multicolumn{1}{|l|}{{L\&A with genre}}  &76.84                \\
\multicolumn{1}{|l|}{{L\&A with emotion}}  &77.66 \\
\multicolumn{1}{|l|}{{L\&A with emotion+genre}}  &\textbf{78.03}    \\

\hline

\end{tabular}
\caption{\label{table:results} classification results based on weighted-F1 score.}
\end{table}
\endgroup

Table \ref{table:results} shows the classification results for predicting the MPAA rating of movies in terms of weighted-F1 score. To disentangle the contributions of genre and emotion vectors to the performance, we experiment with our proposed LSTM with attention architecture without using genre and emotion information (L\&A model). We also investigate the impact of each vector to the results by separately adding them to the model (L\&A with emotion and L\&A with genre). 

The best result is achieved by our proposed ``L\&A with emotion+genre'' model. The weighted F1-score for this model is 78.03\% which is 6.1\% higher than the traditional machine learning model. It also outperforms ``Threshold'' and ``CNN''  baselines by 14.13\% and 3.29\% respectively. Based on the results, both genre and emotion vectors improve the performance of the plain LSTM model with attention. However, the contribution of the genre vector to the model is not as much as the emotions. These results supports our assumption on the relevance of emotion and genre modeling for the task of predicting the MPPA rating. 

\begin{figure}[ht]
\centering
\includegraphics[width=\linewidth]{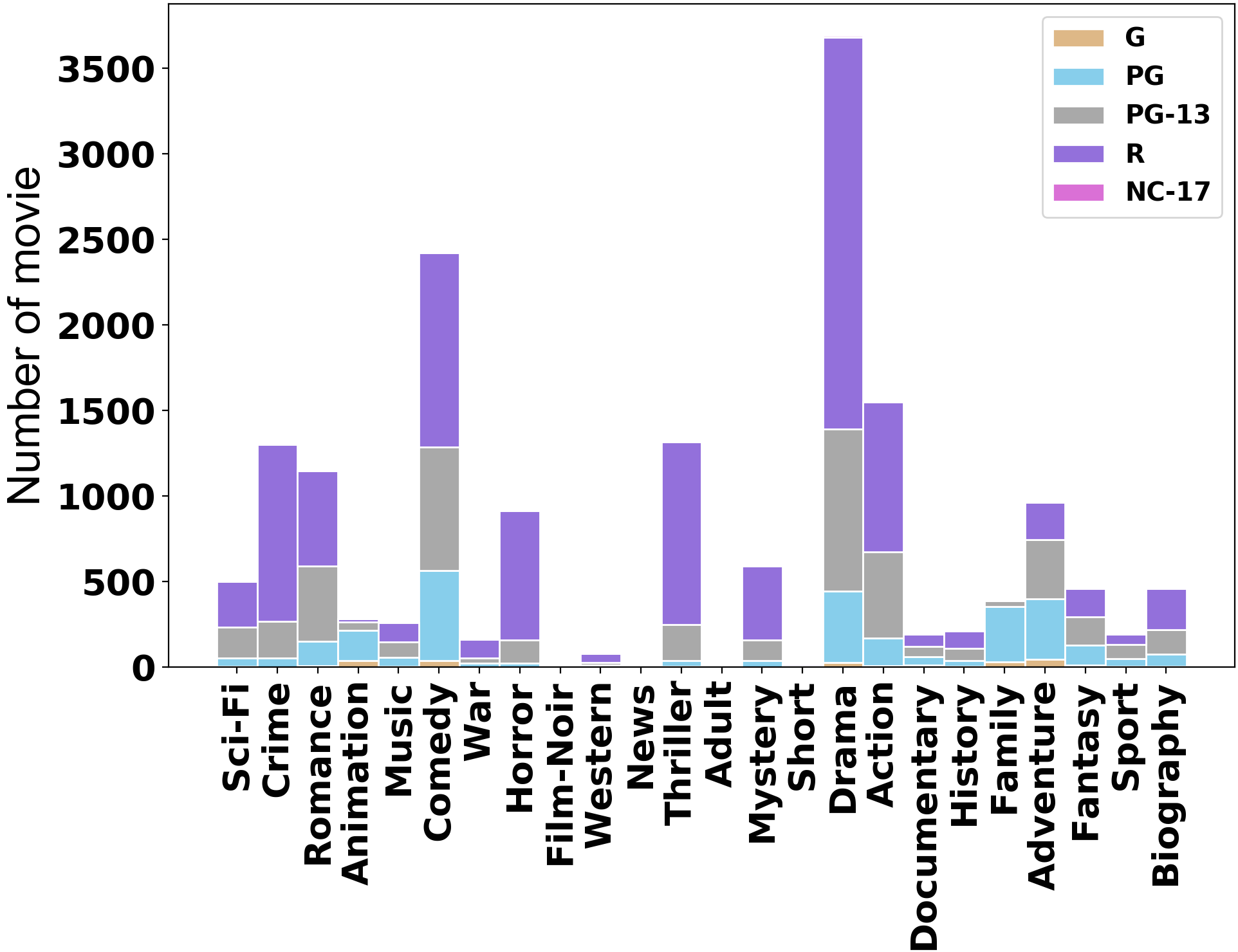}

\caption{MPAA rating distribution per genre.
}

\label{fig:stackBar}
\end{figure}

To shed some light on the effect of genre, we show the distribution of movies in each class for different genres (Figure \ref{fig:stackBar}). According to this figure, some genres are more appropriate for children compared to other genres. For example, MPAA ratings of Animation, Adventure and Family show that most of the movies in these genres are kid-friendly compared to other genres like Drama, Horror, and Crime. 

Although genre can help the model to improve the performance, it is not enough as a single source to predict the MPAA rating. Table \ref{table:genre_based} shows the weighted F1-score for different genres using the our best model. Based on this table, for genres like Comedy and Drama, which contain movies with different MPAA ratings, the performance is as good as Animation, even though most of the Animation movies belong to a single rating. 
For genres like Crime, Horror, and Thriller (that similarly for the most instances have one rating), the performance is better than other genres (84\%,81\%,83\%  respectively), but not that far from genres like Romance and Biography (with 79\% and 86\% respectively) that contain movies with more varied ratings. So, genre by itself is helpful but not the most relevant information to the MPAA rating. 

\begin{table}[ht]
\resizebox{1\columnwidth}{!}{
\centering
\begin{tabular}{|l|c||l|c|}
\hline
\textbf{Genre} &\textbf{F1-score}& \textbf{Genre} &\textbf{F1-score}\\ \hline
{Science-Fiction}  &0.68   & {Action} &0.76               \\
{Family}  &0.70     & {Animation} &0.73                  \\
{Crime} &0.84  & {Biography}  &0.86           \\
{Romance} &0.79  &  {Sport}  &0.79        \\
{Comedy} &0.78     & {Fantasy}  &0.68       \\
{War} &0.74  & {History}  &0.76                \\   
{Horror}  &0.81   & {Documentary}  &0.70          \\
{Adventure}  &0.68  & {Mystery}  &0.79  \\
{Musical}  &0.75   & {Drama}  &0.79       \\
{Thriller}  &0.83   &   {}  &\\
\hline

\end{tabular}
}
\caption{\label{table:genre_based} Weighted F1-score for different genres in the test set }
\end{table}

\begin{figure}[h!]
\centering
\includegraphics[width=1\linewidth]{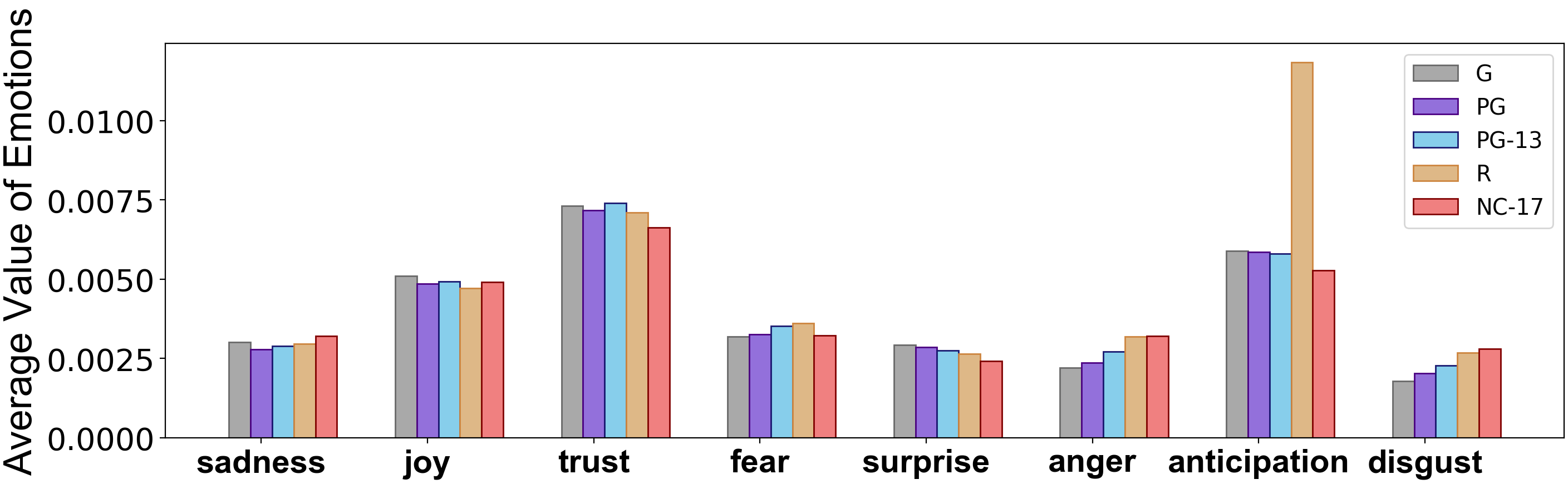}

\caption{MPAA rating distribution per emotion. The y-axis shows the average value for the emotions score for all of the movies in each rating.}
\label{fig:emobar}
\end{figure}

Figure \ref{fig:emobar} shows the average emotion score per each class. According to this figure, some of the emotions are more dominant on a specific class of movies. For example, the values of negative emotions like \textit{disgust}, \textit{anger}, and \textit{sadness} are higher for NC-17 and R compared to the other classes. On the other hand, \textit{surprise} and \textit{trust} show higher rates in G and PG compared to NC-17 and R. These trends bode well with our assumption that emotion vectors could help improving the task.

\section{Error Analysis}
\begin{figure}[h]
\centering
\includegraphics[width=1\linewidth]{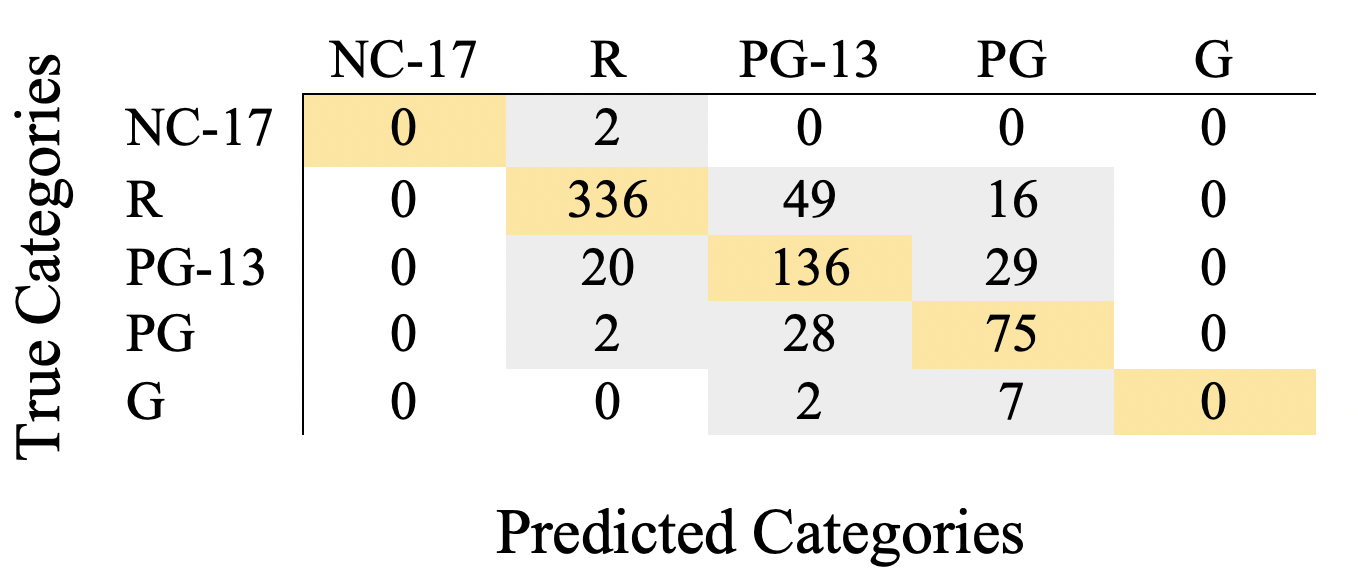}

\caption{Confusion matrix of the best model for predicting MPAA ratings}

\label{fig:conf}
\end{figure}

In order to have a better understanding of the results, we show the confusion matrix of our best model in Figure \ref{fig:conf}. Based on the matrix, our model is not able to correctly predict any instances of G and NC-17. The reason could be the low number of instances in these two classes in our dataset (92 and 14 respectively). On the other hand, the model predicts 83\% of R movies correctly. 75\% of the mistakes in this category are predicted as PG-13, which is the closest group to R.  We plan to investigate these error patterns in our future work.

\subsection{Emotion Analysis}
\begin{figure}[h]
\centering
\subfloat[Average emotion score per rating for correctly classified samples]{%
  \includegraphics[width=\columnwidth]{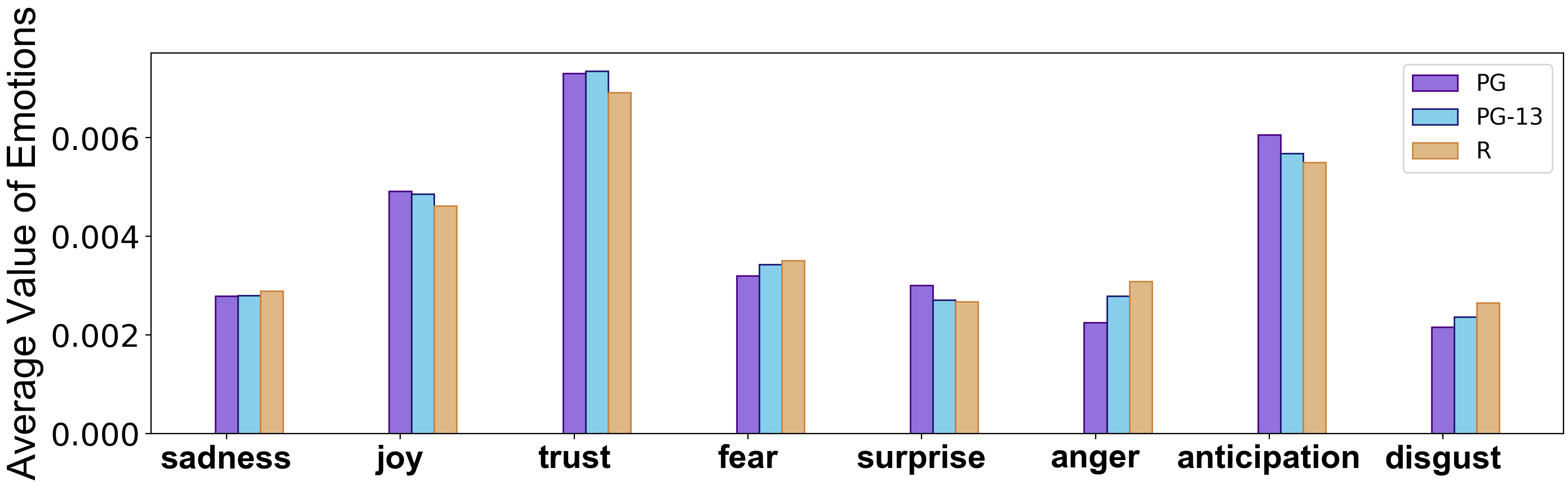}%
  \label{fig:emotionsgroup-a}
}

\subfloat[Average emotion score per rating for miss-classified samples]{%
  \includegraphics[width=\columnwidth]{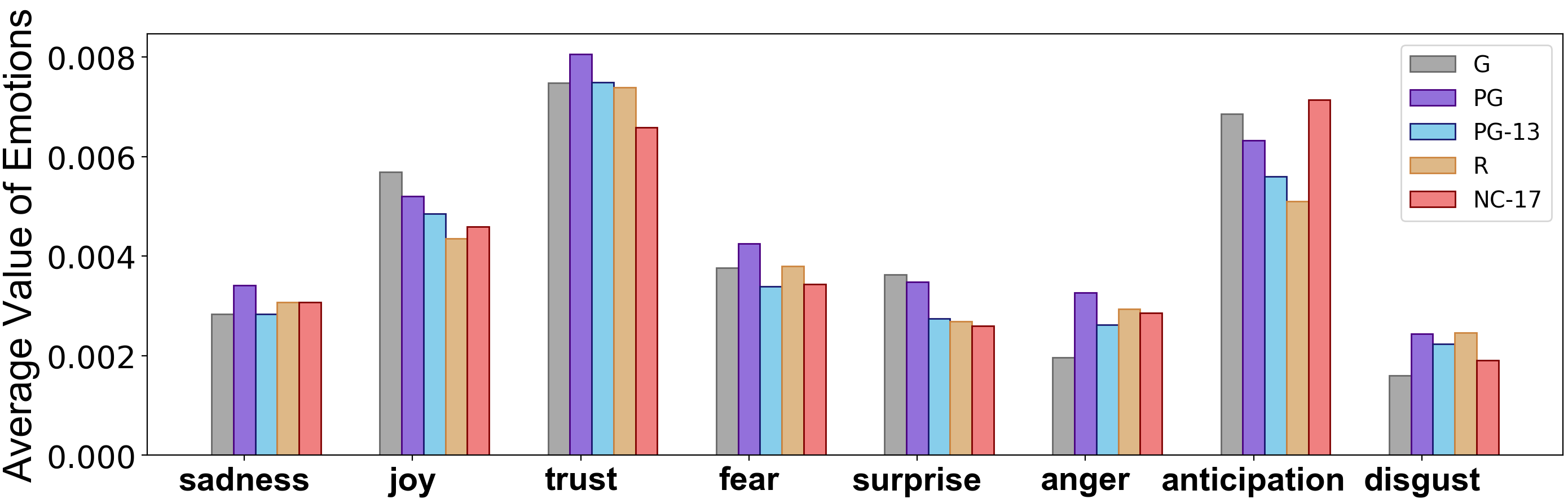}%
  \label{fig:emotionsgroup-b}
}

\caption{Average emotion score of correctly and incorrectly classified movies in the test set per each class.}
\label{fig:emotionsgroup}
\end{figure}

 To further investigate the effects of the emotion vectors, we compare the histogram of emotion scores for correctly and incorrectly labeled instances.
 Figure \ref{fig:emotionsgroup-a} shows an intuitive pattern for average emotion score in samples that are classified correctly (note that we leave G and NC-17 out since we do not have any correct predictions for them). The class R shows the highest rates for the negative emotions. But, in the miss-classified samples \ref{fig:emotionsgroup-b},  we do not observe any clear trends for emotions in different ratings. For example, the PG class shows a high value for negative emotions like disgust, sadness, and anger compared to R, while this is a more appropriate group for children.  
 
 To understand the reason behind this observation, we investigate some samples in the groups PG and R that show high rates of words associated with \textit{disgust}. The results show that those words in R films include terms like \textit{robbery},\textit{murder}, and \textit{asshole}, but in PG rated movies the words associated with \textit{disgust} include \textit{fool}, \textit{sick} and \textit{painful}. Thus, although we have a high number of words in some PG-rated movies that are associated with negative emotions like \textit{disgust}, the degree of negativity, or the strength of the emotion, seems to be lower than in the rated R films ( Table \ref{tab:sample}). And, it seems the model is not able to learn these more subtle differences. 

\begin{table}[]
\resizebox{\columnwidth}{!}{
\begin{tabular}{|l|l|}
\hline
Rate & Sentence                                                 \\ \hline
    & \textcolor{red}{Each victim was killed by a punctuate wound at the skull} \\ 
R    & \textcolor{red}{Goddamn fucking asshole.}                                \\ 
    & \textcolor{red}{Your wife was murdered.}                                 \\ \hline\hline
   & \textcolor{blue}{Lorenz! are you sick?}                                    \\ 
PG   & \textcolor{blue}{Do you think the memory become less painful then!}        \\ 
   & \textcolor{blue}{How terribly awful it all is!}                            \\ \hline
\end{tabular}
}
\caption{Sample sentences of R and PG rated movies that contain ``disgust'' emotion in the conversation. }
\label{tab:sample}
\end{table}

\subsection{Attention Weights}
To analyze the attention weights, we calculate the average weight for all words within each movie script and compare rates for the correctly and incorrectly labeled instances. In other words, for each MPAA rating $r_i$ we calculate the average weights for true positive instances with rating $r_i$ (movies that have rating $r_i$ and predicted as $r_i$), as well as, average weights for a combination of false negative and false positive samples (instances with true rating $r_i$ with prediction $\neq r_i$ and instances predicted as $r_i$ whose true label is $\neq r_i$). 
Table \ref{table:weight2} lists the top 10 words with highest average weights for both groups for the classes PG, PG-13, and R. The fact that for all the classes some bad words are appeared in both \textit{TP} and \textit{FN+FP} categories indicates that the model is biased towards some specific bad words. We plan to employ some methods to reduce this bias through our ongoing research.
\begin{table}[h]
\resizebox{1\columnwidth}{!}{
\centering
\begin{tabular}{|l|l||l|l||l|l|}
    \hline                                                    \multicolumn{2}{|c|}{\textbf{PG}}   &  \multicolumn{2}{c|}{\textbf{PG-13}}    &
    \multicolumn{2}{c|}{\textbf{R}}    
                                                        \\ \hline
\hline
 \textbf{TP} &\textbf{FN+FP}& \textbf{TP} &\textbf{FN+FP} & \textbf{TP} &\textbf{FN+FP}\\ \hline
{bastards}   & {fucked} &  {fucking}   & {australia}  & {australia} &{fucking}     \\

{iced}  & {fuck} &       {fucks}   & {obscure}  & {fucking}&{fuck}     \\

{whore}  & {fucking} &      {fuckin}  & {alexis} &{fuck} & {australia} \\

{shit}  &  {bastards}  &      {australia} & {fuck} &{fucks}& {whore}\\

{penny}    & {fissures}  &     {macram}  & {fuckin}&{fucked}&{fucked }\\

{texts}   & {whore}  &       {niagara}   & {fucking} & {reguiar} &{bastards}    \\ 

{bastard}     & {niagara}  &   {bastards}  & {fissures} &{heshe}&{hookers}\\

{appie}   & {texts}  &    {whore}& {fucked} &{fuckin}&{morn}\\

{peopie}    & {slut}  &      {womanizer}& {niagara}&{temp}&{vader}\\

{morn}     & {arrows}  &     {slut} & {straws}&{kroenen}&{appetite}\\

\hline

\end{tabular}
}
\caption{\label{table:weight2} Top words for the correct and incorrect predicted instances based on average attention weights for PG, PG-13, and R. }
\end{table}

              
                                                    



As our model is not able to predict groups G and NC-17 correctly, we show top 10 words with highest average weights for these groups in a separate table (Table \ref{table:weight}). Although for NC-17, the highest weights are assigned to inappropriate words, the model still cannot predict them correctly. Same for group G, most of the words with high attention are neutral words, but corresponding documents are predicted wrongly. It probably is because of lack of enough samples for these groups.

\begin{table}[h]
\resizebox{1\columnwidth}{!}{
\centering
\begin{tabular}{|l|r|l|r|}
    \hline                                                    \multicolumn{2}{|c|}{\textbf{G}}   &  \multicolumn{2}{c|}{\textbf{NC-17}}         
                                                        \\ \hline
\hline
\textbf{Words} &\textbf{Ave. Weight}& \textbf{Words} &\textbf{Ave. Weight}\\ \hline
{motherfuckin}  &0.061   & {fucking} & 0.054             \\
{texts}  &0.051   & {fuck} &0.028                  \\
{basis} &0.028  & {whore} &  0.024        \\
{crops} &0.027  &  {fucks}  & 0.016       \\
{gang} &0.021     & {fucked}  &0.011       \\
{claus} &0.017  & {drank}  & 0.011               \\   
{initiating}  &0.015   & {cops}  &0.010          \\
{filthy}  &0.015  & {bastards}  &0.010    \\
{catching}  &0.015   & {fuckin}  & 0.008      \\
{instant}  &0.014   & {motherfucker}  & 0.006     \\

\hline

\end{tabular}
}
\caption{\label{table:weight} Top 10 words with highest average weights in miss-classified documents for label G and NC-17.}
\end{table}

\begin{table*}[h]
\centering
\footnotesize
\begin{tabular}{|l|l|l|l|l|}
\hline
\textbf{G}       & \textbf{PG}      & \textbf{PG-13}   & \textbf{R}        & \textbf{NC-17}    \\ \hline
bad (0.034\%)    & bad (0.045\%)    & hell (0.047\%)   & fuck (0.146\%)    & fuckin (0.332\%)  \\ \hline
kill (0.024\%)   & kill (0.022\%)   & shit (0.043\%)   & fucking (0.124\%) & fuck (0.308\%)    \\ \hline
die (0.016\%)    & hell (0.022\%)   & bad (0.041\%)    & shit (0.128\%)    & fucking (0.149\%) \\ \hline
hate (0.014\%)   & die (0.017\%)    & kill (0.033\%)   & hell (0.048\%)    & shit (0.118\%)    \\ \hline
cut (0.014\%)    & cut (0.016\%)    & die (0.023\%)    & kill (0.047\%)    & sex (0.068\%)     \\ \hline
stupid (0.011\%) & hate (0.015\%)   & ass (0.022\%)    & bad (0.044\%)     & cock (0.052\%)    \\ \hline
hell (0.010\%)   & stupid (0.013\%) & cut (0.016\%)    & ass (0.030\%)     & ass (0.054\%)     \\ \hline
hook (0.005\%)   & shit (0.008\%)   & hate (0.015\%)   & bitch (0.028\%)   & bad (0.041\%)     \\ \hline
fat (0.005\%)    & fat (0.006\%)    & stupid (0.014\%) & die (0.025\%)     & kill (0.032\%)    \\ \hline
ugly (0.005\%)   & ass (0.005\%)    & bitch (0.013\%)  & fuckin (0.022\%)  & suck (0.030\%)    \\ \hline
\end{tabular}
\caption{Top 10 bad words in each class. The numbers inside the parenthesis show the ratio of the word across all the scripts of the class~\label{bad-word-ratio}}
\end{table*}

\subsection{Bad Word Ratio}
We conduct an analysis over the same bad word list we used in our ``Threshold'' baseline to further investigate why these words are not enough to predict the MPPA rating of the movies. For each classes in our corpus, we merge all the scripts, and calculate the frequency of bad words over the same class. Table~\ref{bad-word-ratio} shows the top 10 negative words for each classes of data. It is interesting to see that not only the ratio of top bad words' appearance in the scripts, but also the level of negativity of them are different across the classes. That is why the threshold, by its own, is not enough to predict the MPPA ratings. Words like \textit{fat} and \textit{ugly} are mostly listed as offensive words in social media comments, but they are less probable to be considered as inappropriate words in movies. Inversely, we could consider the words like \textit{clit}, \textit{mother-fucker} and \textit{suka} as highly negative, since they are only used in the movie scripts of the classes R and NC-17. These observations bring this idea to the table that as a possible future plan, we can create a lexicon of bad words specified to movies.
 Using our bad word list, we also calculate the overall ratio of negativity per class. The results are shown in Table~\ref{negativity}. Obviously, there is a correlation between the ratio of negativity and the MPPA ratings of movies. 


\begin{table}[h!]
\centering
\footnotesize
\begin{tabular}{|c|c|c|c|c|}
\hline
G                            & PG                          & PG-13                       & R                           & NC-17                       \\ \hline
\multicolumn{1}{|r|}{0.17\%} & \multicolumn{1}{r|}{0.23\%} & \multicolumn{1}{r|}{0.38\%} & \multicolumn{1}{r|}{0.87\%} & \multicolumn{1}{r|}{1.68\%} \\ \hline
\end{tabular}
\caption{Percentage of bad words in different classes~\label{negativity}}
\end{table}

\section{Practicality of MPAA rating in reality}
\begin{figure}[h]
\centering
\includegraphics[width=1\linewidth]{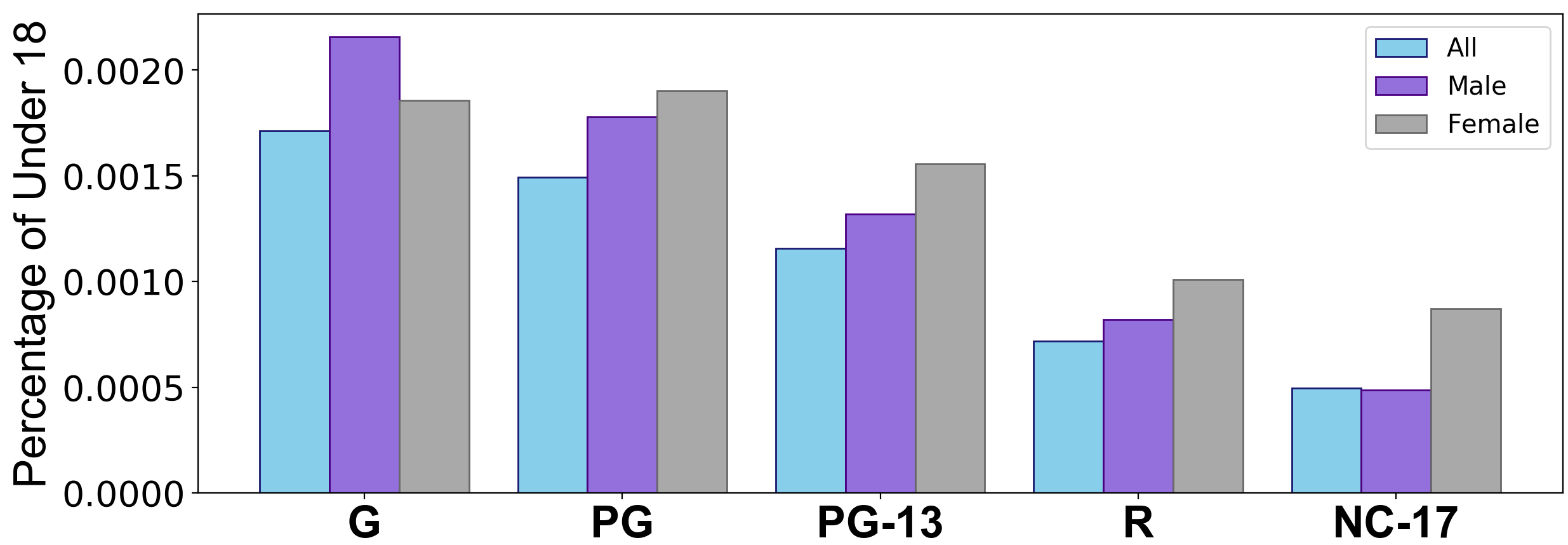}

\caption{Percentage of under 18 audiences in each group of MPAA rating. }

\label{fig:age}
\end{figure}

The whole paper is about predicting MPAA rating, but the important question is that if MPAA ratings could prevent children from watching inappropriate movies or not. There is an element in IMDB website that shows the number of reviewer in different group of ages for each movie. Using this information, we calculate the percentage of reviewer under 18  years old for all movies in the dataset. The result in Figure \ref{fig:age} shows that the more restricted the MPAA rating, the less  under 18 years old people talked about the movie. It possibly means MPAA rating could avoid young people from watching inappropriate movies. 
\section{Conclusion and Future Work}
In this paper, we present a new task of automatic prediction of MPAA rating from movie scripts. We also present a new resource to support the design and benchmarking of the machine learning approaches for the task. Lastly, we propose a neural network architecture to model the conversations among characters of the movies, considering genres and the emotions behind the conversations in order to predict the film rating.

We also show that our best model improves the results compared to the classic machine learning approach, by 6\% weighted F1-score.
As for the future work, we plan to add visual features to our model in order to extract the information that are missing from the video. We also plan to extend the approach to other types of content that are easily accessible to children. 
\bibliography{emnlp-ijcnlp-2019}
\bibliographystyle{acl_natbib}
\end{document}